\documentclass[letterpaper, 12 pt, conference]{ieeeconf}  

\IEEEoverridecommandlockouts                              
\usepackage{graphicx} 
\usepackage{bm}

\usepackage{amsmath}
\usepackage{amssymb}
\usepackage{mathtools}
\usepackage{multirow}
\usepackage{booktabs}
\title{\LARGE \bf
ED-VAE: Entropy Decomposition of ELBO in Variational Autoencoders}

\author{Fotios Lygerakis$^{1}$ and Elmar Rueckert$^{1}$
\thanks{}
\thanks{$^{1}$Chair of Cyber-Physical Systems, University of Leoben, Austria}%
}

\begin{document}

\maketitle
\thispagestyle{empty}
\pagestyle{empty}

\begin{abstract}

Traditional Variational Autoencoders (VAEs) are constrained by the limitations of the Evidence Lower Bound (ELBO) formulation, particularly when utilizing simplistic, non-analytic, or unknown prior distributions. These limitations inhibit the VAE’s ability to generate high-quality samples and provide clear, interpretable latent representations. This work introduces the Entropy Decomposed Variational Autoencoder (ED-VAE), a novel re-formulation of the ELBO that explicitly includes entropy and cross-entropy components. This reformulation significantly enhances model flexibility, allowing for the integration of complex and non-standard priors. By providing more detailed control over the encoding and regularization of latent spaces, ED-VAE not only improves interpretability but also effectively captures the complex interactions between latent variables and observed data, thus leading to better generative performance.

\end{abstract}

\section{Introduction}
\label{sec:intro}
Variational Autoencoders (VAEs) \cite{vae} have emerged as a compelling framework for generative modeling, offering a principled approach to learning probabilistic representations of complex data distributions. At the heart of VAEs lies the ELBO, which serves as the objective function guiding the optimization of the model parameters. The ELBO intertwines reconstruction accuracy with a regularization term that encourages the learned latent variables to adhere to a specified prior distribution, typically a standard normal distribution. This dual objective underpins the VAE's capacity to generate new, plausible data samples while uncovering interpretable and disentangled representations of the underlying data-generating process.
\begin{figure}[!h]
\centering
  \includegraphics[width=0.8\linewidth]{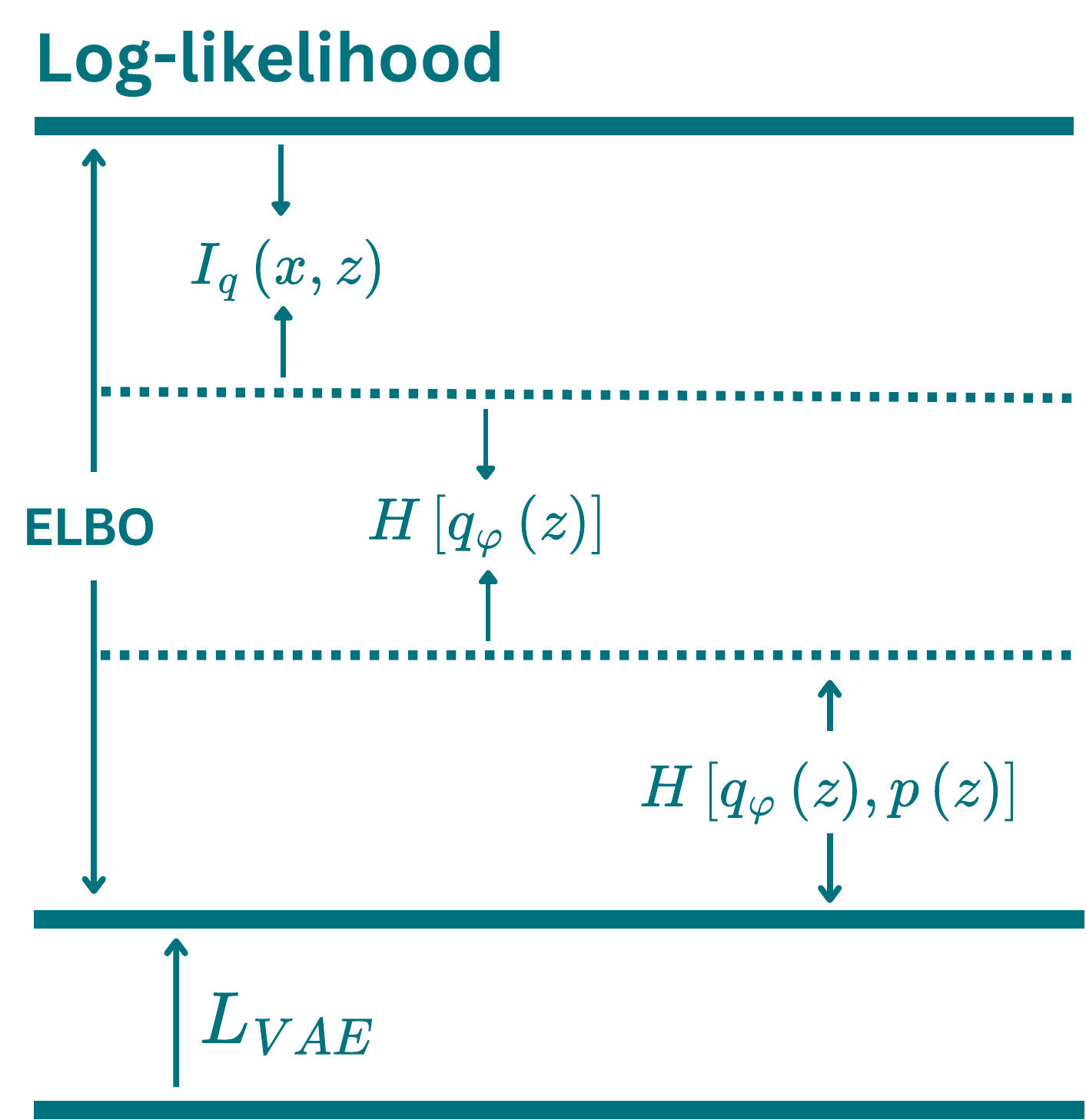}  
    \caption{Visual description of the Evidence Lower BOund (ELBO) in a ED-VAE. Figure~\ref{fig:ed_vae_graph} describes the minimization of ELBO, which we show consists of a mutual information, an entropy and a cross-entropy term, as shown in Section~\ref{sec:entropy_dec}.}
    \label{fig:ed_vae_graph}
\end{figure}
However, the conventional formulation of the ELBO presents certain limitations that may hinder the flexibility and interpretability of VAEs \cite{elbo_surgery}. Primarily, the choice of the prior distribution in the ELBO plays a critical role in shaping the learned latent space, yet the traditional formulation does not offer an intuitive understanding of how the chosen prior interacts with the learned latent distribution. Moreover, the standard ELBO does not explicitly account for the amount of information shared between the data and the latent variables, which can be instrumental in understanding and controlling the representational power of the VAE \cite{CriticVAE, lygerakis2023crvae}. This formulation often leads to additional limitations in model flexibility and interpretability, particularly when the prior distributions cannot be captured as normal Gaussians or when the chosen priors do not have an analytic form for computing the components of the ELBO. Additionally, the traditional ELBO does not effectively handle situations where the prior distribution is unknown or where it cannot be easily integrated due to its complexity. This restricts the VAE’s ability to adequately capture the intricate interactions between latent variables and observed data, thereby hindering the model's capacity to generate high-quality, diverse samples and provide interpretable and disentangled representations of the underlying data-generating process.

Motivated by these challenges, we propose a novel reformulation of the ELBO that seeks to enhance the expressivity, interpretability, and adaptability of VAEs. Our formulation decomposes the ELBO into distinct terms that separately account for reconstruction accuracy, mutual information between data and latent variables, and a marginal KL divergence to prior. 
We further decompose the marginal KL divergence term into entropy and cross-entropy terms, which facilitates the more explicit control of the inherent uncertainty in the latent variables and the alignment of the learned latent distribution with the prior. 
Our proposed Entropy Decomposed VAE (ED-VAE) offers, thus, a more granular perspective on the latent variable modeling in VAEs, allowing for a deeper understanding of the intricate balance between data reconstruction, information preservation, and regularization. 
The explicit depiction of entropy and cross-entropy terms enables a more informed and flexible choice of priors, extending the exploration beyond standard normal priors to incorporate domain-specific knowledge or desirable properties into the latent space. 
Furthermore, the delineation of mutual information facilitates direct control over the expressiveness and diversity of the latent space, which is crucial for enhancing the generative capabilities of the VAE. 
In our experiments, we compared the performance of the traditional VAE with our entropy-decomposed variation using two synthetic datasets. These datasets were created to test how each model handles priors of varying complexities, from simple Gaussian to complex non-Gaussian distributions. 

\section{Related Work}
The development of VAEs \cite{vae} has seen significant advancements, particularly in the formulation and optimization of the ELBO. Foundational work by \cite{elbo_surgery} introduced the concept of dissecting the ELBO into more interpretable components, which allows for an improved understanding of how different elements influence training and inference in VAEs. 

Similarly, \cite{infoVAE} addressed the imbalance in traditional ELBO formulations in their development of InfoVAE, which often prioritizes either reconstruction accuracy or latent regularization at the expense of the other.

Building on the need for robustness in ELBO computation, especially with complex and non-analytic priors, the utilization of advanced sampling methods has proven crucial. \cite{pmlr-v151-kviman22a} explored the use of multiple-importance sampling within the ELBO, improving the accuracy and stability of the variational inference. 
Their approach to deep ensembles for variational approximations presents a novel way to handle uncertainties in model predictions. 

The issues of model robustness and flexibility are further compounded by challenges such as posterior collapse, which affects the quality of generated samples and the diversity of the latent space. \cite{DBLP:journals/corr/abs-1911-02469} provided an in-depth analysis of posterior collapse.
Additionally, the work on learning optimal priors for task-invariant representations by \cite{10.1145/3534678.3539291} underscores the importance of flexible and adaptive prior settings in VAEs.
Approaches like the CR-VAE \cite{lygerakis2023crvae}, which uses contrastive regularization to prevent posterior collapse ensure a robust representation in the latent space and contribute to a more robust and adaptive generative modeling framework.

\section{Method}

In this section, we present a novel formulation of the ELBO used in VAEs, which enhances interpretability and flexibility in choosing the latent variable prior. We start from the original ELBO formulation, introduce an intermediate form that includes a mutual information term and a KL divergence to the prior term, and finally propose a new form of ELBO that decomposes the later KL divergence into entropy and cross-entropy terms.

\subsection{Original ELBO Formulation}

The original ELBO is formulated as:
\begin{equation}
    L(\theta; \phi) = \mathbb{E}_{q_{\phi}(z|x)}[\log p_{\theta}(x|z)] - KL(q_{\phi}(z|x) || p(z))
\end{equation}
Table \ref{tab:naming} contains the description of all the terms in this and the following forms of the ELBO.
Following the analysis in \cite{elbo_surgery}, the ELBO can be reformulated to make the mutual information term explicit:
\begin{equation}
\begin{split}
        & L(\theta; \phi) = \mathbb{E}_{q_{\phi}(z|x)}[\log p_{\theta}(x|z)] - \log N  \\
        & + \mathbb{E}_{q_\phi(z)}[H[q_\phi(n|z)]] - KL(q_\phi(z) || p(z))
\end{split}
\label{eq:info_elbo}\end{equation}

where: $H[q_\phi(n|z)]$ is the entropy of the conditional distribution and $\log N$ and $\mathbb{E}_{q_\phi(z)}[H[q_\phi(n|z)]]$ together form the mutual information term. Consequently Equation \ref{eq:info_elbo} takes the form
\begin{equation}
    \begin{split}
        & L(\theta; \phi) = \mathbb{E}_{q_{\phi}(z|x)}[\log p_{\theta}(x|z)] - I_{q}(x, z) \\
        & - KL(q_\phi(z) || p(z))
    \end{split}
        \label{eq:info_elbo_2}
\end{equation}

\subsection{Entropy Decomposed ELBO}
\label{sec:entropy_dec}
We further decompose the marginal KL divergence to the prior into an entropy and a cross-entropy term as:
\begin{equation}
    \begin{aligned}
    & KL(q_\phi(z) || p(z)) \\
    &= \mathbb{E}_{q_\phi(z)} \left[ \log q_\phi(z) - \log p(z) \right] \\
    &= \mathbb{E}_{q_\phi(z)} \left[ \log q_\phi(z) \right] - \mathbb{E}_{q_\phi(z)} \left[ \log p(z) \right] \\
    &= - H[q_\phi(z)] + H[q_\phi(z), p(z)]
    \end{aligned}
\end{equation}
We thus propose a new form of ELBO by substituting the above decomposed KL divergence in \ref{eq:info_elbo_2}:
\begin{equation}
    \begin{aligned}
    & L(\theta; \phi) = \mathbb{E}_{q_{\phi}(z|x)}[\log p_{\theta}(x|z)] - I_{q}(x, z) \\
    & + H[q_\phi(z)] - H[q_\phi(z), p(z)] 
    \end{aligned}    
\end{equation}

This novel formulation of the ELBO, through the decomposition of the KL divergence into entropy and cross-entropy terms, unveils a more granular perspective of the latent variable modeling in VAEs. It fosters a deeper understanding of the interplay between the latent variable posterior and the chosen prior, thereby allowing for more informed and flexible prior selections.

Firstly, the explicit representation of the entropy term, $H[q_\phi(z)]$, reveals the inherent uncertainty or randomness in the encoded latent variables. This facilitates a more direct analysis and control over the latent space's expressiveness and diversity, which is crucial for the generative capabilities of the VAE.

Secondly, the cross-entropy term, $H[q_\phi(z), p(z)] $, acts as a clear measure of alignment or divergence between the learned latent distribution and the predefined prior. This explicit depiction allows for a more intuitive and adaptable choice of priors, enabling the exploration beyond standard normal priors, and allowing the incorporation of domain-specific knowledge or desirable properties into the latent space. It is important to state here that unlike the KL divergence, where choosing a prior that allows for an analytic form is crucial, using the cross-entropy term only requires the ability to sample from the considered prior distribution.
\begin{table}\centering
\caption{Naming of referenced notation.}\label{tab:naming}
\begin{tabular}{ll}
\toprule
$p_{\theta}(x|z)$ & likelihood \\
$q_{\phi}(z|x)$ & approximate posterior \\
$p(z|x)$ & true posterior \\
$p(z)$ & prior \\
$q_{\phi}(z)$ & encoding distribution\\
$I_{q}(x, z)$ & mutual information of $x$ and $z$\\
$D_{KL}(q_\phi(z|x)||p(z))$ & KL divergence to prior\\
$D_{KL}(q_\phi(z)||p(z))$ & marginal KL divergence to prior\\
\bottomrule
\end{tabular}
\end{table}
\subsection{ELBO Optimization}

We adopt a Gaussian distribution assumption for the likelihood \( p_{\theta}(x|z) \), simplifying the reconstruction loss to the Mean Squared Error (MSE) between the original and reconstructed data (\ref{appendix_1}). This choice is congruent with the continuous nature of the datasets, ensuring well-behaved gradients for backpropagation.

\begin{equation}
    \mathcal{L}_{\text{recon}} = \mathbb{E}_{q_{\phi}(z|x)}\left[ \|x - \hat{x}\|^2 \right] \nonumber
\end{equation}

For optimizing the mutual information term, we employ the InfoNCE loss, providing a lower bound to it. In the work of \cite{CPC}, it has been established that the InfoNCE loss, \(\mathcal{L}_{\text{InfoNCE}}\), serves as a lower bound to the mutual information, \(I_q(x,z)\):
\begin{equation}
    I_q(x,z) \geq \log(K) - \mathcal{L}_{\text{InfoNCE}} \nonumber
\end{equation}
where \(K\) represents the batch size. $\mathcal{L}_{\text{InfoNCE}}$ is computed utilizing positive and negative sample pairs, encouraging the encoder to generate representations that are more informative of the data.

\begin{equation}
    \mathcal{L}_{\text{InfoNCE}} = -\log \frac{\exp(z_i^T z_j)}{\sum_{k=1}^K \exp(z_i^T z_k)} \nonumber
\end{equation}

Similarly, the entropy $\mathcal{L}_{\text{Ent}}$ and cross-entropy $\mathcal{L}_{\text{XEnt}}$ terms are computed using samples from the batch to facilitate their estimation.

The total loss of ED-VAE is
\begin{equation}
    \mathcal{L}_{\text{ED-VAE}}(\theta; \phi) = \mathcal{L}_{\text{recon}} - \mathcal{L}_{\text{InfoNCE}} - \mathcal{L}_{\text{Ent}} +   \mathcal{L}_{\text{XEnt}}
\end{equation}
All the sub-losses are jointly optimized through backpropagation, with gradients propagated through the encoder and decoder parameters. 

\section{Experiments}

Our experimental design evaluates the traditional VAE and the ED-VAE using synthetic datasets designed to highlight different complexities in data distributions. These experiments are specifically tailored to assess the models' ability to handle varying complexities of priors, from Gaussian to non-Gaussian complex distributions.

\subsection{Synthetic datasets}

We utilize two distinct synthetic datasets to evaluate the performance of the traditional VAE and ED-VAE:

\paragraph{Dataset 1: Gaussian Prior} This dataset consists of data generated from a multi-dimensional Gaussian distribution. The latent variables \(z\) are sampled from a standard Gaussian distribution \( \mathcal{N}(0, 1)\), and the observed data is created by a linear transformation of \(z\) followed by Gaussian noise. Additionally, positive samples are generated by adding small Gaussian noise around the original data points, simulating slight variations within the data distribution that are still characteristic of the Gaussian prior. This serves as a baseline, testing the models under standard conditions where the data aligns well with the models' Gaussian prior assumptions.

\paragraph{Dataset 2: Complex Non-Gaussian Prior} This dataset features a complex structured prior modeled as a mixture of Gaussians modulated by sinusoidal functions. The latent variables are sampled from multiple Gaussian distributions with varying means and scales, modulated by a sinusoidal function to introduce non-linear interactions within the latent space. The observed data is again generated by a linear transformation followed by Gaussian noise, with positive samples created similarly to Dataset 1. This dataset tests the standard VAE’s capability against the ED-VAE's flexibility in handling complex, structured non-Gaussian distributions.

\subsection{Model Configurations}

The traditional VAE is configured with a standard Gaussian prior, which is expected to perform adequately on Dataset 1 but may struggle with Dataset 2 due to its simplicity. In contrast, the ED-VAE is set up with a flexible prior configuration for Dataset 2, aiming to closely approximate the complex distributions involved, while maintaining a similar setup as Dataset 1 for direct comparison. We evaluate the two models based on the ELBO of held-out data, which consists of the lower bound for the log-likelihood of the data distribution learned.

\subsection{Experimental Setup}

To ensure robust and reliable findings, each experiment is conducted five times with different seeds, and results are averaged to offset the effects of random initialization and stochastic optimization. Both models consist of a two-layer fully connected multilayer perception, with 400 hidden dimensions and a five-dimensional latent space, with the decoder following the inverse architecture. The VAEs are trained using the Adam optimizer with a learning rate of 0.001 and a batch size of 512. Training continues for 1000 epochs with early stopping based on validation loss to prevent overfitting. Input features are normalized to prevent bias.

The computation of the entropy and cross-entropy terms, critical for the ED-VAE model, is handled distinctively to ensure accurate representation and alignment with the model's theoretical foundations. For the traditional VAE, the entropy term is derived directly from the latent variable's log-variance output by the encoder, representing the inherent uncertainty of the encoded representations. Specifically, entropy is calculated using the formula:
\begin{equation}
H[q_{\phi}(z)] = 0.5 \sum(1 + \log(\sigma^2)) \nonumber
\end{equation}

where \(\sigma^2\) is the variance of the latent variables.

For the ED-VAE, the cross-entropy term \( H[q_{\phi}(z), p(z)] \) is computed to measure how well the encoded latent distribution aligns with the specified prior. This involves evaluating the negative log-likelihood of the latent variables under the chosen prior distribution, using a Gaussian mixture model (GMM) \cite{gmm} when non-standard complex priors are employed. The GMM parameters are estimated from the prior data, and the cross-entropy is calculated as the expected negative log-likelihood under this model. When the prior is a standard Gaussian, the cross-entropy simplifies evaluating the Gaussian density function at the values of the encoded latent variables.

\begin{table}[t]
\caption{Comparative performance metrics for VAE and ED-VAE models on two datasets with distinct prior distributions. Metrics include Mean Squared Error (MSE), Kullback-Leibler Divergence (KLD), and Evidence Lower Bound (ELBO), averaged over five trials. Lower values indicate better performance for MSE and KLD, whereas higher values are better for ELBO, computed as $ELBO = -MSE - KLD$.}

\label{tab:model_performance}
\vskip 0.15in
\begin{center}
\begin{small}
\begin{sc}
\begin{tabular}{lcccc}
\toprule
Metric & Model & Dataset 1 & Dataset 2 \\
\midrule
\multirow{2}{*}{MSE} & VAE & 2.78 $\pm$ 0.1 & 20.7 $\pm$ 0.5 \\
                     & ED-VAE & \textbf{1.54 $\pm$ 0.05} & \textbf{19.36 $\pm$ 0.3} \\
\midrule
\multirow{2}{*}{KLD} & VAE & 10.2 $\pm$ 0.2 & 12.5 $\pm$ 0.4 \\
                     & ED-VAE & \textbf{0.0 $\pm$ 0.0} & \textbf{0.05 $\pm$ 0.01} \\
\midrule
\multirow{2}{*}{ELBO} & VAE & -12.98 $\pm$ 0.3 & -33.2 $\pm$ 0.6 \\
                      & ED-VAE & \textbf{-1.54 $\pm$ 0.02} & \textbf{-19.41 $\pm$ 0.2} \\
\bottomrule
\end{tabular}
\end{sc}
\end{small}
\end{center}
\vskip -0.1in
\end{table}

\section{Results}

The evaluation of two VAEs, the traditional one and the entropy-decomposed one, was conducted across two datasets designed to test model efficacy under different prior assumptions: a standard Normal prior and a complex, non-Gaussian prior.

On the dataset with the standard Normal prior, both models performed competently (Table \ref{tab:model_performance}), yet the ED-VAE showed a distinct advantage in its encoding and regularization capabilities. This was evident from its more efficient data representation and notably superior regularization, leading to a higher ELBO. The traditional VAE, while effective, demonstrated less optimal alignment with the normal prior, indicative of its comparatively limited capacity to regulate the latent space.

The differences between the models became more pronounced when faced with the complex, structured non-Gaussian prior. Here (Table \ref{tab:model_performance}), the traditional VAE struggled to adapt, reflecting its challenges with modeling a more complex prior distribution with a normal one. 
Conversely, the ED-VAE managed to maintain a much higher level of data fidelity (higher ELBO) and minimal divergence from the complex prior. This performance underscores the ED-VAE's robust adaptability and its effective management of the latent space to align closely with even highly irregular priors.

The superior performance of the ED-VAE across both datasets underscores its effectiveness in managing complex data distributions. The proposed objective for training the VAE, incorporating entropy and cross-entropy components, allows for better control over latent space regularization. This not only results in better alignment with the priors but also enhances the interpretability and the quality of the data reconstructions.

\section{Discussion}

In this work, we presented a novel approach to training VAEs by decomposing the ELBO formulation into an entropy and cross-entropy term. Our approach showcases better adaptability and performance on two synthetic datasets with variable complexity of prior distributions. This is achieved through a reformulated ELBO that explicitly accounts for the entropy in the latent variables and their alignment with respective priors, enhancing the model's ability to handle intricate data distributions.

Our findings emphasize the necessity of selecting suitable prior models for generative tasks, particularly highlighted by the traditional VAE's difficulties with complex data that deviates from normal distributions. The ED-VAE, with its flexible framework, enables a nuanced interaction between the model and the underlying characteristics of the data, leading to improved learning outcomes and more precise reconstructions.

Despite the benefits, the introduction of entropy and cross-entropy terms does increase computational demands, particularly affecting memory usage and processing power. Addressing these challenges will be essential to optimize the model's efficiency and facilitate its wider adoption.

In the future, we plan to apply the ED-VAE to real-world image datasets that present complex distributional characteristics and refine its architecture to enhance its efficacy. Additionally, we aim to develop methods for computing the cross-entropy term without knowing the dataset's prior, using techniques like unsupervised learning to infer the prior distribution, enabling the ED-VAE's use in scenarios with undefined priors.





\bibliography{main}
\bibliographystyle{ieeetr}

\newpage
\appendix
\onecolumn
\section{Supplementary Material}

\subsection{Analysis of the Cross-Entropy Term with a Standard Normal Prior}
\label{appendix_1}
Given a standard normal prior $p(z) = \mathcal{N}(0, I)$, the probability density function is expressed as 
$p(z) = \frac{1}{\sqrt{2\pi}}e^{-\frac{1}{2}z^2}$. The log of this probability density function simplifies to 
$\log p(z) = -\frac{1}{2}\log(2\pi) - \frac{1}{2}z^2$.

The cross-entropy term is defined as the expectation of the negative log-likelihood of the latent variables 
encoded by $q_\phi(z)$ under the standard normal prior:

\begin{equation}
H[q_\phi(z), p(z)] = -\mathbb{E}_{q_\phi(z)}[\log p(z)]
\end{equation}

Substituting the log probability density of $p(z)$ into the equation, we get:

\begin{equation}
H[q_\phi(z), p(z)] = -\mathbb{E}_{q_\phi(z)}\left[-\frac{1}{2}\log(2\pi) - \frac{1}{2}z^2\right] = \frac{1}{2}\mathbb{E}_{q_\phi(z)}[\log(2\pi) + z^2] = \frac{1}{2}\log(2\pi) + \frac{1}{2}\mathbb{E}_{q_\phi(z)}[z^2]
\end{equation}

Here, $\log(2\pi)$ is a constant and can be taken out of the expectation. The term $\mathbb{E}_{q_\phi(z)}[z^2]$ 
represents the expected squared norm of the latent variables under the distribution $q_\phi(z)$. By minimizing 
the cross-entropy term, the encoded latent variables are encouraged to align with the characteristics of the 
standard normal prior, particularly having a unit average squared distance from the origin.

\section{Generation of Complex Non-Gaussian Data with Positives}

\subsection{Latent Variable Generation}
The latent variables \( z \) are generated from a non-Gaussian prior, specifically a modulated mixture of Gaussians. The generation process involves the following steps:

\begin{enumerate}
    \item \textbf{Mixture Components}: The latent variables are drawn from multiple Gaussian distributions, where each component \( i \) of the mixture has a mean \( \mu_i \) and scale \( \sigma_i \). The means and scales are linearly spaced between specified bounds:
    \[
    \mu_i = \text{linspace}(-3, 3, 3)
    \]
    \[
    \sigma_i = \text{linspace}(0.5, 1.0, 3)
    \]

    \item \textbf{Latent Sampling}: For each component, samples are drawn as follows:
    \[
    z[i] = \mathcal{N}(\mu_i, \sigma_i^2)
    \]

    \item \textbf{Modulation}: Each component is modulated by a sinusoidal function to introduce non-linear interactions:
    \[
    z_{\text{modulated}}[i] = z[i] + \sin(0.5 \pi z[i])
    \]

    \item \textbf{Dimension Matching}: If the desired latent dimensionality \( \text{latent\_dim} \) is greater than the number of components, additional Gaussian noise is added. If \( \text{latent\_dim} \) is less, the dimensions are truncated.
\end{enumerate}

\subsection{Data Generation}
Observable data points \( x \) are generated using a linear transformation of the latent variables followed by the addition of Gaussian noise:

\begin{enumerate}
    \item \textbf{Transformation Matrix}: A transformation matrix \( W \) is sampled:
    \[
    W \sim \mathcal{N}(0, 1), \quad W \in \mathbb{R}^{\text{latent\_dim} \times \text{data\_dim}}
    \]

    \item \textbf{Data Construction}: The data points are constructed as:
    \[
    x = z_{\text{modulated}} W + \mathcal{N}(0, 0.5)
    \]
\end{enumerate}

\subsection{Positive Sample Generation}
Positive samples are generated by adding Gaussian noise to simulate slight variations within the data distribution:
\[
x_{\text{positive}} = x + \mathcal{N}(0, \text{radius})
\]
where \(\text{radius}\) determines the variability of the positives from the anchor data points.

\end{document}